\documentclass{article}

\usepackage[preprint]{corl_2026} 

\usepackage{amsmath}
\usepackage{enumitem}
\usepackage{graphicx}
\usepackage{booktabs}
\usepackage{graphicx}
\usepackage{pdflscape}
\usepackage{array}
\usepackage{longtable}
\usepackage[dvipsnames]{xcolor}

\title{RECALL: Recovery Experience Collection for Active Lifelong Learning in Vision-Language-Action Models}

\author{
  Ulas Berk Karli\\
  Department of Computer Science\\
  Yale University 
  United States\\
  \texttt{ulasberk.karli@yale.edu} \\
  \And
  Tesca Fitzgerald \\
  Department of Computer Science\\
  Yale University 
  United States\\
  \texttt{tesca.fitzgerald@yale.edu} \\
}

\begin{document}
\maketitle


\begin{abstract}
    Vision-Language-Action (VLA) models are commonly fine-tuned through passive imitation learning, where additional demonstrations are collected for tasks where the policy performs poorly. This approach incurs several downsides: it requires the robot to fail before data collection is triggered, provides little guidance about which states require supervision, and wastes demonstrator effort on redundant parts of the task where the policy already performs well. In this paper, we propose an active, continual learning paradigm for VLAs. We demonstrate that active, uncertainty-guided data collection leads to more efficient fine-tuning than when using passively-collected demonstrations. However, we also find that fine-tuning only on actively-collected recovery data leads to catastrophic forgetting. We evaluate techniques for continual learning, including replay-based data mixing and elastic weight consolidation, and identify tradeoffs between plasticity to uncertainty-guided recovery data and retention of previously learned behaviors. Overall, our work contributes an empirical study of active continual learning for autoregressive VLAs, establishing that uncertainty-guided recovery demonstrations can improve adaptation efficiency while also revealing open challenges when targeted new data is incorporated into large robot policies.
    
\end{abstract}



\section{Introduction}
Recent Vision-Language-Action (VLA) models such as RT-1~\cite{rt1}, RT-2~\cite{rt2}, Gemini Robotics~\cite{gemini},  $\pi_0$-FAST~\cite{fast}, and $\pi_{0.5}$~\citep{pi05} have shown that large-scale robot learning can produce broadly capable manipulation policies. However, despite their generality, VLAs still require fine-tuning when deployed to new robots, environments, or task distributions. In practice, this adaptation is usually performed through passive imitation learning, where an expert (i) collects additional demonstrations starting from task initial states, (ii) fine-tunes the model, (iii) evaluates performance, and (iv) repeats if the policy remains unreliable.
This process is inefficient for several reasons. First, it only reveals that the model needs more data after failures occurred, which can be time-consuming and unsafe. Second, it gives little guidance about how much additional data is needed, which can lead to under- or over-collection of data. Third, passive recollection does not identify where supervision would be most useful. Demonstrators may therefore spend most of their efforts on states that are already well represented in the original training data; yet, failures may arise from a small set of difficult substeps or recovery from distribution-shifted intermediate states.

We propose an active, continual learning paradigm for fine-tuning VLAs. Our central hypothesis is that collecting demonstrations from high-uncertainty states will lead to more informative (and thus, efficient) data collection compared to typical passive data collection. We use a state-of-the-art technique for uncertainty quantification~\citep{karli2025insightinferencetimesequenceintrospection} and utilize it to 
collect data that more directly addresses the failure modes limiting downstream task success.
We then evaluate techniques for continual learning, with the aim of fine-tuning a VLA with uncertainty-guided data \emph{without} catastrophic forgetting of previously-learned tasks. Our contributions are as follows:

\begin{enumerate}[leftmargin=*]
    \item An active data collection framework for autoregressive VLAs that uses INSIGHT~\citep{karli2025insightinferencetimesequenceintrospection} to select uncertain states to prompt new demonstrations and provide the way to best utilize that data.
    \item An evaluation of active vs passive data collection, showing that demonstrations provided from high-uncertainty states lead to better model performance than unguided demonstrations.    
    \item A comparison of online- and offline-style data collection, showing that collecting data from the first uncertain state in a rollout is just as effective (and thus, more efficient) compared to collecting data from all uncertain states.
    \item Application and analysis of continual-learning techniques (regularization- and replay-based) for VLA models, showing that replay-based data mixing is currently the most reliable strategy.
    \item Empirical design lessons and open research questions for continual, active learning in VLAs.
\end{enumerate}

    
\section{Related Works}


\textbf{Active Learning for Robot Policies. }
Active learning involves identifying states, trajectories, tasks, or environments where additional demonstrations would be most informative. Interactive imitation learning methods such as DAgger~\cite{dagger} and HG-DAgger~\cite{hg-dagger} reduce covariate shift by collecting supervision on states induced by the learned policy. Uncertainty-guided active learning further targets this process by querying the expert only when the policy is uncertain, for example using Monte Carlo dropout to select states for supervision~\citep{cui2019uncertaintyawaredataaggregationdeep}. However, most uncertainty-aware aggregation methods \citep{menda2019ensembledaggerbayesianapproachsafe, zhao2025conformalizedinteractiveimitationlearning, Wang24, lee2025diffdaggeruncertaintyestimationdiffusion} have been studied for smaller, task-specific policies, often in driving~\citep{hg-dagger, Wang24} or classical imitation-learning settings~\citep{dagger}, rather than for large pretrained VLAs.

\textbf{Uncertainty Quantification in VLAs. }
Prior work on uncertainty quantification (UQ) has explored ensemble methods~\citep{lakshminarayanan2017simplescalablepredictiveuncertainty}, Bayesian approximations~\citep{gal16dropout}, conformal prediction~\citep{angelopoulos2022cp}, and learned intervention or failure predictors. For example, KnowNo~\citep{knowno2023} uses conformal prediction to decide when a VLM planner should ask for help, but operates over high-level plan rather than low-level VLA actions.  INSIGHT~\citep{karli2025insightinferencetimesequenceintrospection} is an inference-time introspection method for UQ in autoregressive VLAs that uses token-level probability features to predict when the robot should ask for help. INSIGHT learns a model from token-level uncertainty features extracted during action decoding, including entropy, log-probability, and Dirichlet-based aleatoric and epistemic uncertainty estimates. 

\textbf{Continual Learning. }
Continual learning studies how models acquire new knowledge without \emph{catastrophic forgetting}, where training on new data degrades performance on earlier tasks or distributions~\citep{MCCLOSKEY1989}. This issue has been studied in LLMs~\citep{luo2025empiricalstudycatastrophicforgetting}, suggesting that large pretrained sequence models such as VLAs may be similarly vulnerable when fine-tuning on new data. Common solutions include replay-based methods, which mix new data with examples from prior datasets~\citep{rebuffi2017icarlincrementalclassifierrepresentation,lopezpaz2017}, and regularization-based methods, which constrain changes to parameters important for prior behavior. Elastic weight consolidation (EWC) is a representative regularization method that uses a Fisher information matrix to penalize changes to parameters important for previous tasks~\citep{Kirkpatrick_2017_ewc}. 
\label{sec:background}

\begin{figure}[t]
    \centering
    \includegraphics[width=\linewidth]{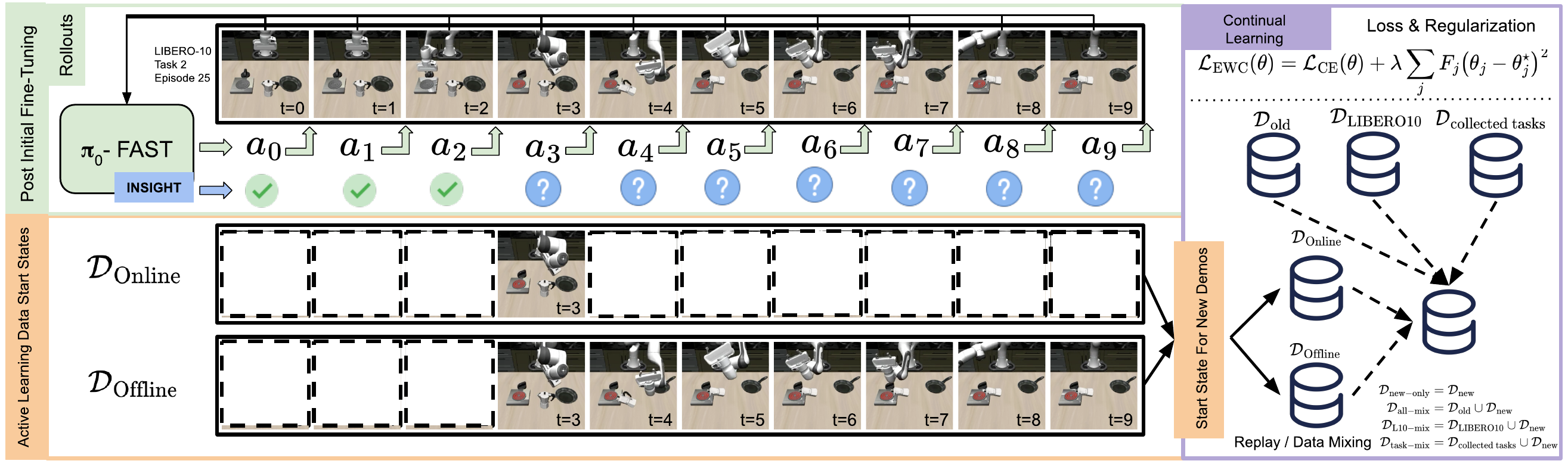}
    \caption{
\textbf{Overview of uncertainty-guided active continual learning.}
An initial $\pi_0$-FAST policy is rolled out on LIBERO-10 while INSIGHT identifies high-uncertainty states. We compare passive collection with online recovery from the first high-uncertainty state and offline recovery from all high-uncertainty states, then integrate the resulting data using replay, or EWC regularization.
}
    \label{fig:overview_pipeline}
\end{figure}


\section{Enabling Active, Continual Learning from Uncertainty-Guided Data}

Active learning in VLAs remains unsolved and under-explored. In this work, we propose an active, continual learning paradigm for fine-tuning VLAs. We focus on autoregressive VLAs, particularly $\pi_0$-FAST~\citep{fast}, which tokenizes continuous robot actions and predicts them via a next-token objective.

\textbf{Toward enabling active learning} in VLAs, we leverage INSIGHT for uncertainty quantification~\citep{karli2025insightinferencetimesequenceintrospection} and utilize it by interpreting its ``help predictions'' as a proxy for high-uncertainty states from which to collect new demonstrations. 
We study whether uncertainty-guided data collection enables effective active learning, leading to our first research question \textbf{RQ-1:} Compared to passive data collection, do demonstrations originating from high-uncertainty states result in better policy performance?

We next consider the effect of online vs offline active learning in VLA performance. In online learning, the robot requests a demonstration in real-time, which originates at the \emph{first} uncertain state within a rollout. In contrast, an offline learning setting involves the robot performing a full rollout, recording \emph{every} uncertain state to prompt for demonstrations at a later. We pose \textbf{RQ-2:} Is online-style data collection sufficient, or do we need dense offline collection from all uncertain states?

\textbf{Toward \emph{continual} active learning} for VLAs, we evaluate \textbf{RQ-3:} Can autoregressive VLAs be fine-tuned \emph{only} on newly collected recovery data? This is in contrast to the dataset aggregation assumed for RQ-1\&2, which combine the robot's previous dataset and the new uncertainty-guided data.
Finally, we apply and evaluate standard techniques for continual learning to address \textbf{RQ-4:} Do data replay and/or regularization methods mitigate catastrophic forgetting while still enabling adaptation to new training data?

To evaluate these RQs, we propose a pipeline (Figure~\ref{fig:overview_pipeline}) consisting of an active learning component (to identify uncertain states to prompt new data collection) and a continual learning component (fine-tuning the model on newly-collected data without degrading prior capabilities).

\subsection{Active Learning Pipeline}

We begin with a $\pi_0$-FAST policy trained on LIBERO~\citep{liu2023liberobenchmarkingknowledgetransfer} via the $\pi_0$-FAST authors' recipe for 30k steps, using the final checkpoint as the initial policy $\pi_{\theta_0}$. $\pi_0$-FAST tokenizes continuous action chunks and trains with a next-token prediction objective. Given language instruction $\ell$, visual observation $o_t$, and robot state $s_t$, the policy predicts a future action-token chunk. This policy serves both as the LIBERO-10 baseline and as the rollout policy used to induce states for active data collection. 

During each rollout $\tau_i$, we apply INSIGHT~\citep{karli2025insightinferencetimesequenceintrospection} at each policy step to create a set of candidate recovery timesteps $\mathcal{T}_i=\{\forall t \in \tau_i \; | \; H(o_t, s_t, \ell, \pi_{\theta_0})=1\}$ where $H_t=1$ indicates that INSIGHT predicts the policy requires assistance in that state. We evaluate \textit{Strong INSIGHT}, which is trained with step-wise labels indicating when the policy requires help during real robot executions, and \textit{Weak INSIGHT}, which is trained from episode-level success/failure labels on the LIBERO dataset. These terms describe the model training strategy, not downstream task difficulty.  
While Weak INSIGHT is more aligned with the LIBERO tasks in our evaluation, we prioritize Strong INSIGHT throughout our experiments due to its generalization across real-world and simulated settings (as shown in \cite{karli2025insightinferencetimesequenceintrospection}).

We consider two active recovery datasets. The \textit{online} dataset collects one recovery demonstration from the first high-uncertainty state in each rollout, $t_i^{\mathrm{online}}=\min\{t:h_t=1\}$, simulating deployment-time intervention. The \textit{offline} dataset collects from every high-uncertainty state, $\mathcal{T}_i^{\mathrm{offline}}=\mathcal{T}_i$, simulating offline data collection from completed rollouts. We also collect passive start-state baselines that match the task distribution and number of demonstrations of the online and offline datasets, isolating the value of collecting from high-uncertainty states rather than merely adding more task-relevant data.
To collect new demonstration data, we reset the simulator to each state within each dataset and record an expert recovery trajectory to task completion, forming datasets $\mathcal{D}_{\mathrm{offline}}$, $\mathcal{D}_{\mathrm{online}}$, and $\mathcal{D}_{\mathrm{passive}}$.
%
%
After data collection, we fine-tune with the same autoregressive cross-entropy objective used for the initial policy, $\mathcal{L}_{\mathrm{CE}}(\theta)=-\sum_{t=1}^{T}\log p_\theta(x_t\mid x_{<t})$, where $x_{1:T}$ includes prompt, observation-conditioned context, and action tokens. 

\subsection{Continual Learning Strategies}


\textbf{Replay-based data mixtures. }
Let $\mathcal{D}_{\mathrm{new}}$ denote the newly collected recovery dataset (either $\mathcal{D}_{\mathrm{offline}}$, $\mathcal{D}_{\mathrm{online}}$, or $\mathcal{D}_{\mathrm{passive}}$). 
We compare new-only fine-tuning on $\mathcal{D}_{\mathrm{new}}$, full replay on $\mathcal{D}_{\mathrm{old}}\cup\mathcal{D}_{\mathrm{new}}$, LIBERO-10 replay on $\mathcal{D}_{\mathrm{LIBERO10}}\cup\mathcal{D}_{\mathrm{new}}$, and targeted replay on $\mathcal{D}_{\mathrm{collected\ tasks}}\cup\mathcal{D}_{\mathrm{new}}$. Colllected tasks refer to the tasks determined to have low accuracy and used in active learning based data collection. These mixtures test whether recovery data alone is sufficient, whether full prior-data replay is necessary, and whether replay restricted to targeted tasks preserves non-collected tasks.

\textbf{Elastic weight consolidation (EWC). }
We evaluate EWC as a regularization-based continual-learning method. EWC penalizes movement away from the initial parameters $\theta_0$ using a diagonal Fisher estimate $F$ computed on a reference dataset: $\mathcal{L}_{\mathrm{EWC}}(\theta)=\mathcal{L}_{\mathrm{CE}}(\theta)+\lambda\sum_j F_j(\theta_j-\theta_{0,j})^2$. The coefficient $\lambda$ controls the stability-plasticity tradeoff: larger values better preserve parameters important for prior behavior, but may reduce adaptation to recovery demonstrations. 

\textbf{Learning-rate ablations. }
Finally, we test whether smaller updates alone reduce forgetting. Standard fine-tuning uses the original learning-rate schedule from the baseline VLA training setup, consisting of a linear warmup to $\alpha=2.5\times10^{-5}$ followed by cosine decay to a high learning rate of $\alpha=2.5\times10^{-6}$. We compare this against a low constant learning rate, $\alpha=2.5\times10^{-8}$, separating the effect of reduced update magnitude from explicit continual-learning regularization.
\label{sec:approach}



	


\section{Experiment Overview and General Setup}
\label{sec:experiments_results}
We perform 5 experiments, corresponding to our 4 research questions and 1 follow-up experiment. Each experimental involves configuring the following independent variables:
\begin{itemize}[leftmargin=*]
    \item \textbf{State Selection:} \underline{passive} start-state collection vs active learning (guided either by \underline{Strong INSIGHT} or \underline{Weak INSIGHT}, and producing either \underline {online} or \underline{offline} datasets). 
    \item \textbf{Training Data Buffer:} \underline{full replay} vs \underline{new-only} vs \underline{LIBERO-10 replay} vs \underline{targeted replay}
    \item \textbf{Training Loss Function: } Standard cross-entropy \underline{(CE)} vs \underline{EWC} regularization
    \item \textbf{Learning Rate: } \underline{Standard} vs low, \underline{constant LR}
\end{itemize}

Each experiment involves evaluating a set of policies (corresponding to various experimental conditions) with 50 rollouts per task in LIBERO-10, for a total of 500 rollouts at each checkpoint. Unless otherwise stated, replay-based active-learning experiments use new normalization statistics (recomputed on the exact dataset being used to train the model). Appendix~\ref{app:additional_results} provides best-checkpoint summaries, pairwise statistical tests, normalization-statistics ablations, and additional EWC sweeps. 

We report three metrics: \textit{overall} success across all 10 tasks, \textit{collected-task} success on the 5 tasks used for recovery-data collection, and \textit{retained-task} success on the remaining 5 tasks. In all bar plots, statistical comparisons are computed with two-sided two-proportion $z$-tests over rollout success counts. Significance markers indicate $p<0.05$ (*), $<0.01$ (**), and $<0.001$ (***). In all line graphs, we denote baseline performance (dashed line) and its 95\% Wilson CI (shaded band). 

\section{Experiment 1: Active Collection vs. Passive Start-State Collection}
\label{sec:exp_active_vs_passive}

We address \textbf{RQ-1:} compared to passive data collection, do demonstrations from high-uncertainty states lead to better policy performance? We compare performance after fine-tuning the baseline VLA on \underline{passive} data vs active, \underline{online} data collected via either Strong or Weak INSIGHT. We train the model on a full replay buffer using the standard LR schedule and CE loss. 

\begin{figure}
  \centering
  \begin{minipage}[b]{0.48\textwidth}
  \includegraphics[width=\linewidth]{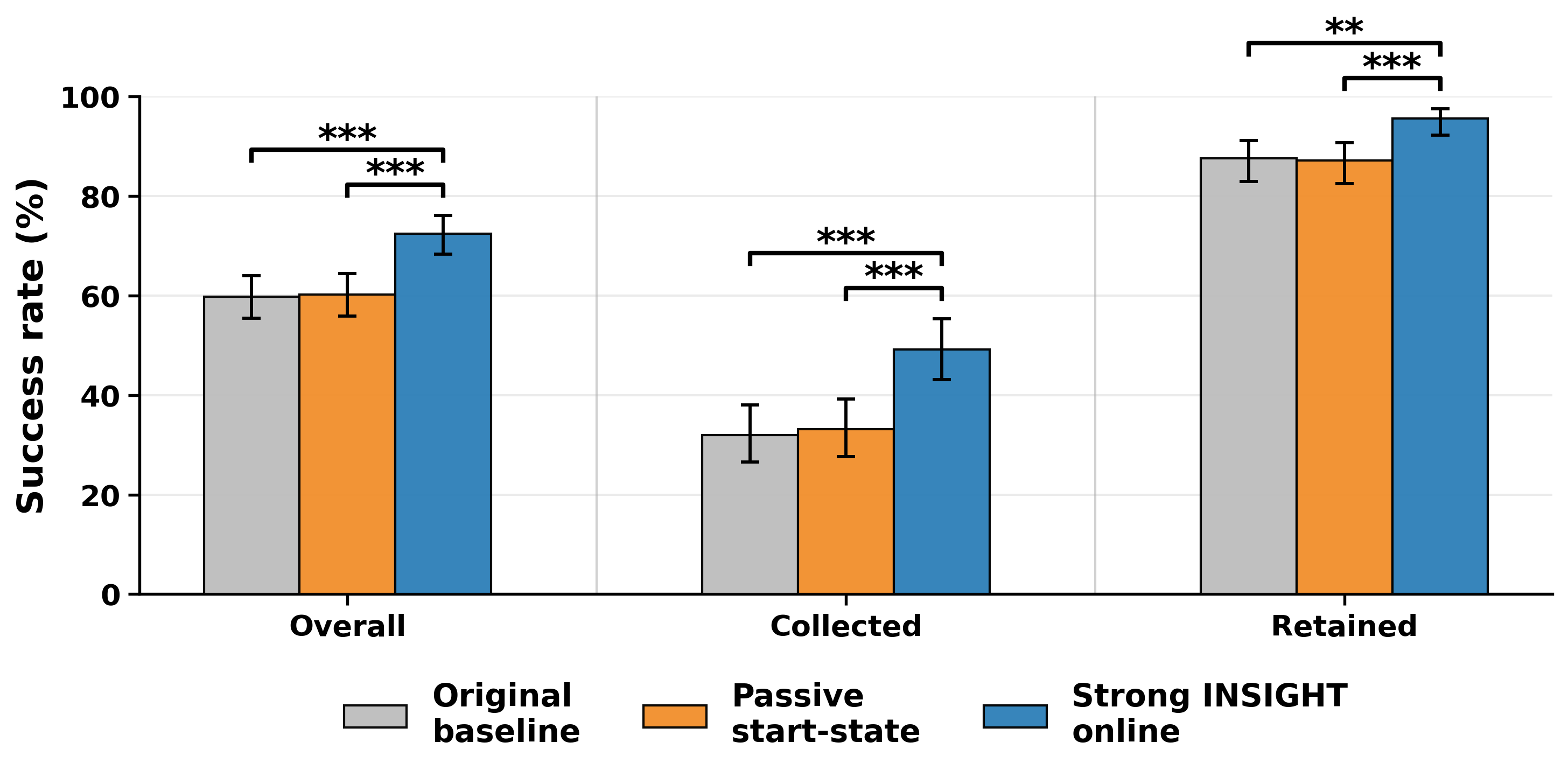}
  \label{fig:rq1_active_vs_passive_strong}
  \end{minipage}
  \hfill
  \begin{minipage}[b]{0.48\textwidth}
  \includegraphics[width=\linewidth]{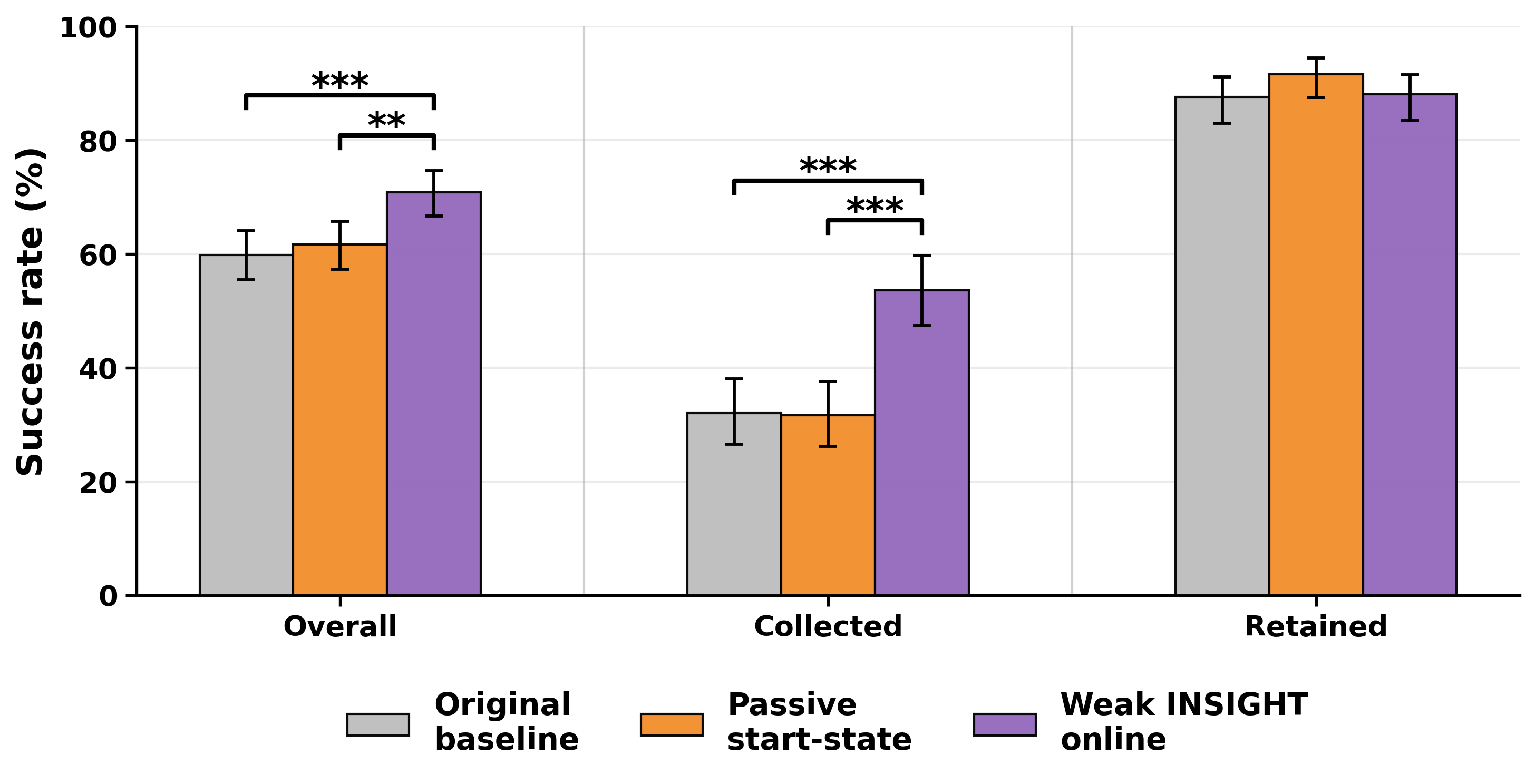}
  \label{fig:rq1_active_vs_passive_weak_new_norm}
  \end{minipage}
  \caption{
  \textbf{Online recovery collection improves over passive start-state collection.}
  We compare fine-tuned policy performance after adding matched passive or active data. Active collection is guided by Strong INSIGHT (left) or Weak INSIGHT (right).
  }
  \label{fig:rq1_active_vs_passive}
\end{figure}

\textbf{Results: }
Figure~\ref{fig:rq1_active_vs_passive} shows that online recovery collection outperforms matched passive collection for both INSIGHT variants. In the Strong INSIGHT setting, online recovery reaches 72.4\% overall success, compared to 60.2\% for matched passive collection and 59.8\% for the original baseline. This improvement over passive collection is statistically significant ($p=0.0001$). The Weak INSIGHT comparison shows the same qualitative trend under the same new-normalization setting.

\textbf{Discussion: }
These results show that the benefit of active collection is not simply due to adding more demonstrations from the same tasks. When task distribution and data volume are controlled, demonstrations from high-uncertainty states provide more useful supervision than demonstrations from task initial states. Appendix~\ref{app:norm_stats_ablation} reports old-normalization active-versus-passive comparisons for both Strong and Weak INSIGHT. These runs show more muted improvement, consistent with old normalization statistics acting as a stabilizing constraint that can preserve prior behavior but limit adaptation to the recovery-data distribution. This motivates our use of old normalization statistics in the new-only experiments, where we test recovery-data-only adaptation under a conservative setting.

\section{Experiment 2: Online vs. Offline Recovery Collection}
\label{sec:exp_online_vs_offline}

We address \textbf{RQ-2:} is online-style collection from the first high-uncertainty state sufficient, or is offline collection from all high-uncertainty states necessary? We compare performance using \underline{online} vs \underline{offline} datasets. Both datasets are informed by Strong INSIGHT, and use a full replay buffer and the standard CE loss and LR schedule.

\begin{figure}[b]
  \centering
  \begin{minipage}[b]{0.6\textwidth}
    \includegraphics[width=\linewidth]{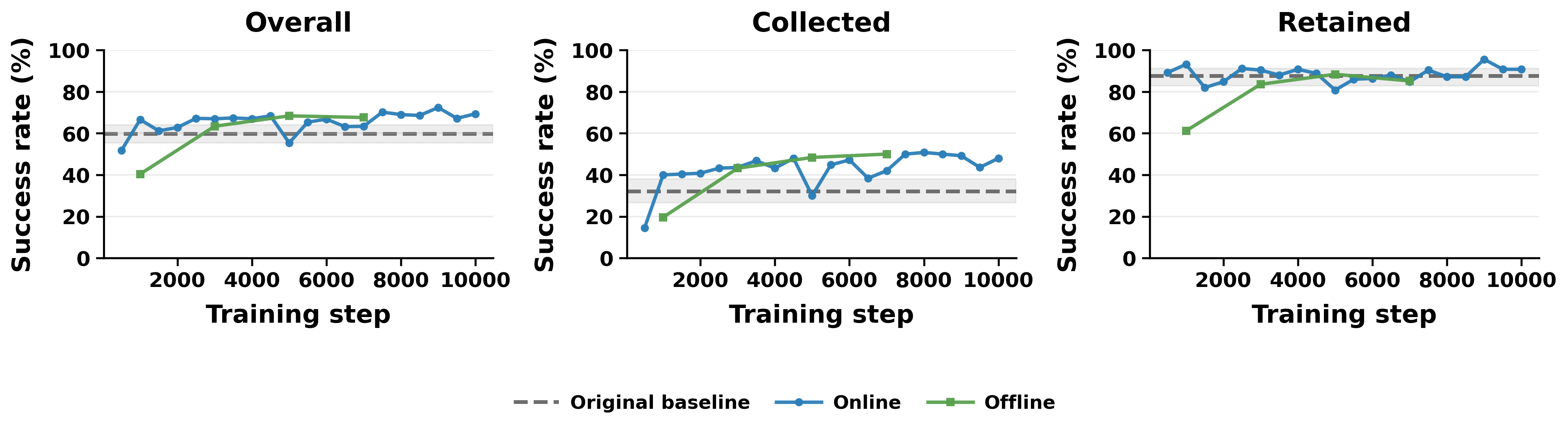}
  \end{minipage}
  \hfill
  \begin{minipage}[b]{0.38\textwidth}
    \includegraphics[width=\linewidth]{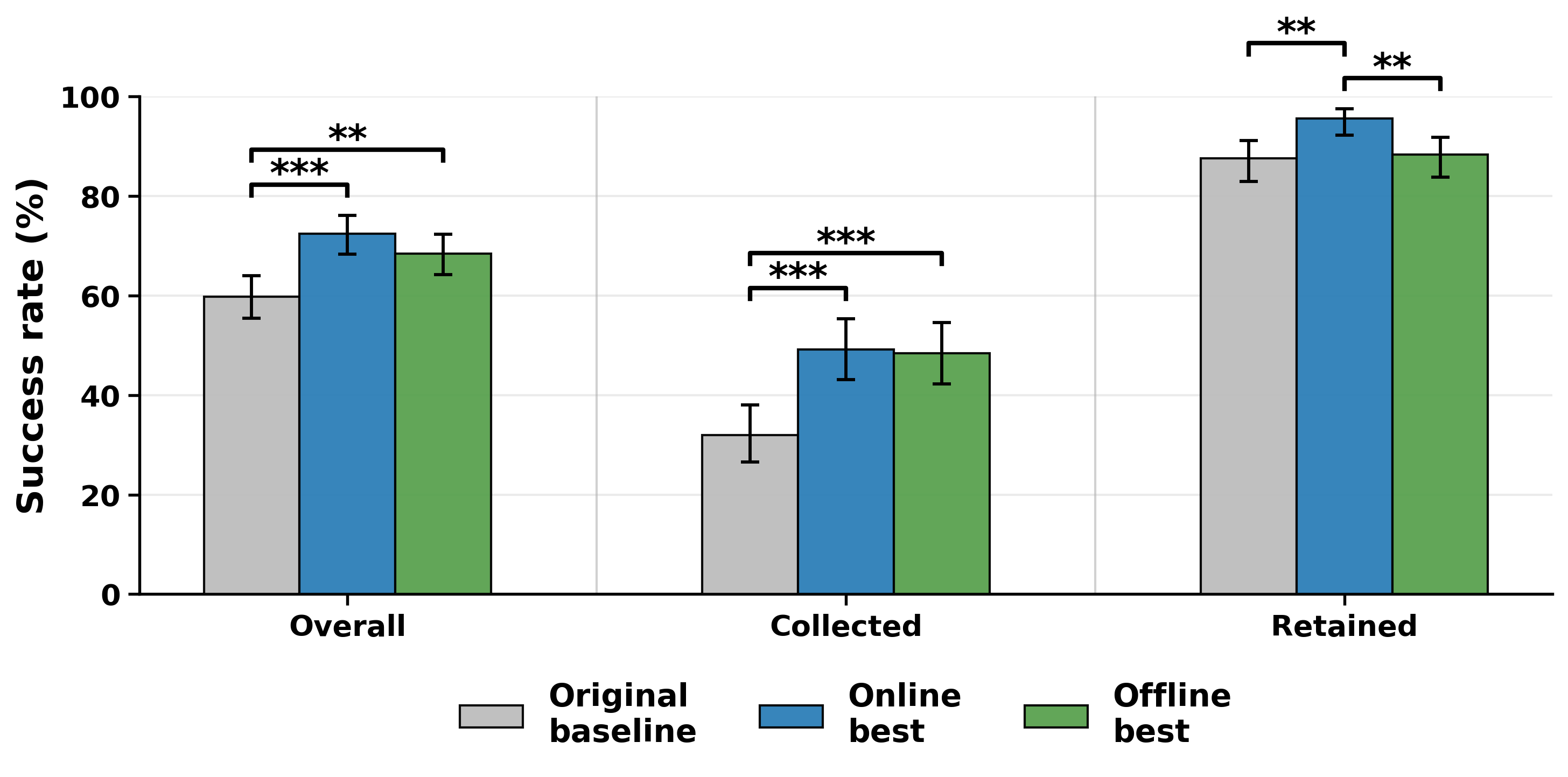}
  \end{minipage}
  \caption{
  \textbf{Online recovery collection is as effective as offline collection despite using fewer demonstrations.}
  Left: training curves for online and offline Strong INSIGHT recovery datasets mixed with prior demonstrations. Right: best-checkpoint comparison across overall, collected-task, and retained-task success.
  }
  \label{fig:rq2_online_vs_offline}
\end{figure}

\textbf{Results: }
Figure~\ref{fig:rq2_online_vs_offline} compares online and offline recovery collection. Online collection reaches 72.4\% overall success at its best checkpoint, while offline collection reaches 68.4\%. The two methods achieve nearly identical collected-task performance, with online collection reaching 49.2\% and offline collection reaching 48.4\%. The overall difference is not statistically significant ($p=0.1659$).

\textbf{Discussion: }
We interpret this as a demonstration-efficiency result rather than strict superiority. Online collection reaches comparable performance while requiring fewer recovery demonstrations. This suggests that the first reliable high-uncertainty state may mark the point where the rollout begins to diverge from successful behavior, so a corrective demonstration from that state can supervise both the uncertain state and the remainder of the task. Dense offline collection adds more data, but those additional states do not translate into proportionally better adaptation.

\section{Experiment 3: New-Only Fine-Tuning}
\label{sec:exp_new_only}

We address \textbf{RQ-3:} can autoregressive VLAs be fine-tuned only on newly collected recovery data? We compare performance after fine-tuning on \underline{full replay} data vs \underline{new-only} data. New data is collected using Strong INSIGHT online, Strong INSIGHT offline, and Weak INSIGHT online. We reuse old normalization statistics from the original LIBERO dataset for new-only training as a regularizer: old norms partially constrain distribution shift, so collapse under this setting provides stronger evidence that recovery data alone is insufficient.

\begin{figure}[h]
    \centering
    \includegraphics[width=\linewidth]{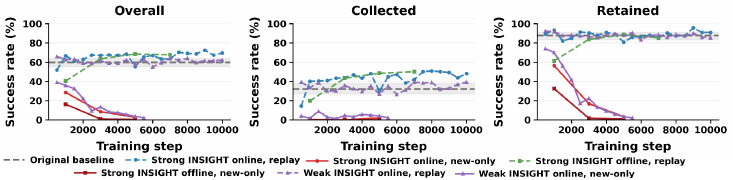}
    \caption{
    \textbf{New-only fine-tuning causes catastrophic forgetting.}
    We compare training on new-only data (solid lines) against replay-based training that combines old and new data (dashed lines).
    }
    \label{fig:rq3_new_only_forgetting}
\end{figure}

\textbf{Results: }
Figure~\ref{fig:rq3_new_only_forgetting} shows that new-only fine-tuning causes severe forgetting. Across Strong and Weak INSIGHT recovery datasets, retained-task performance collapses as training progresses and overall success falls far below the original baseline. In contrast, replay-based all-mix training preserves retained-task performance while improving collected tasks. For example, the best Strong INSIGHT online new-only checkpoint reaches only 28.4\% overall success, with 0.4\% collected-task success and 56.4\% retained-task success. The corresponding replay-based online model reaches 72.4\% overall success, with 49.2\% collected-task success and 95.6\% retained-task success.

\textbf{Discussion: }
Recovery demonstrations are informative precisely because they focus on difficult, policy-induced states, but this also makes the dataset distributionally narrow. New-only fine-tuning does not provide enough coverage to preserve behaviors that the original policy already performs well. Thus, active learning for VLAs cannot be reduced to selecting high-value states; the adaptation procedure must also preserve prior competence.

\section{Experiment 4: Continual-Learning Regularization}
\label{sec:exp_ewc}

We address \textbf{RQ-4:} do learning-rate reduction and regularization mitigate catastrophic forgetting while still enabling adaptation to recovery data? Using new-only data, we compare \underline{CE with standard LR} vs \underline{CE with constant LR} vs \underline{EWC} with constant LR and multiple coefficient values. Full replay results are included as a reference.

\begin{figure}
    \centering
    \includegraphics[width=\linewidth]{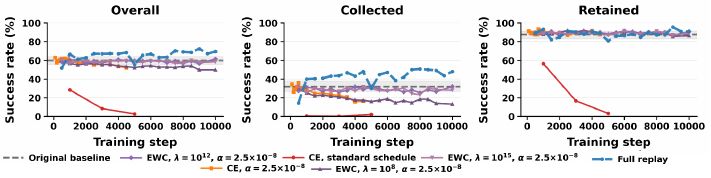}
    \caption{
    \textbf{Low learning rates ($\alpha$) and EWC reduce forgetting but limit adaptation to new data.}
    Compared to full replay, low learning-rate fine-tuning and EWC better preserve retained-task performance than standard new-only fine-tuning, but produce smaller gains on collected tasks.
    }
    \label{fig:rq4_ewc_low_lr_tradeoff}
\end{figure}

\textbf{Results: }
Figure~\ref{fig:rq4_ewc_low_lr_tradeoff} compares representative regularization settings against low-learning-rate fine-tuning and replay. Low-learning-rate new-only training reaches 62.8\% overall success and preserves retained-task performance at 91.2\%, but collected-task success remains only 34.4\%. EWC with $\lambda=10^{12}$ reaches 61.4\% overall success, with 32.0\% collected-task success and 90.8\% retained-task success. These methods preserve prior behavior better than standard new-only fine-tuning, but do not match full replay, which reaches 72.4\% overall success and 49.2\% collected-task success.

The full EWC coefficient sweep and filtered-Fisher sweep are reported in Appendix~\ref{app:ewc_sweeps}. They show the same pattern: stronger regularization can keep retained-task success near the original baseline, but collected-task performance remains limited and overall success does not improve. Filtering the Fisher reference data changes the stability-plasticity behavior but does not remove the core tradeoff.

\textbf{Discussion: }
Low learning rates and EWC solve only part of the active continual-learning problem. Weak constraints allow more adaptation but risk forgetting, while strong constraints preserve retained-task behavior but suppress useful learning from recovery demonstrations. In our setting, replay provides a stronger stability-plasticity tradeoff than parameter regularization alone.

\section{Experiment 5: Replay Scope}
\label{sec:exp_replay_scope}

Motivated by the effectiveness of replay, we investigate how much prior data should be replayed. Using Strong INSIGHT online recovery data, we compare \underline{full replay} vs \underline{LIBERO-10 replay} vs \underline{targeted replay}(containing the subset of LIBERO-10 replay data relevant to the collected tasks ). We also investigate a \underline{filtered EWC} variant where the Fisher matrix excludes the collected tasks. 

\begin{figure}[b]
  \centering
  \begin{minipage}[b]{0.6\textwidth}
    \includegraphics[width=\linewidth]{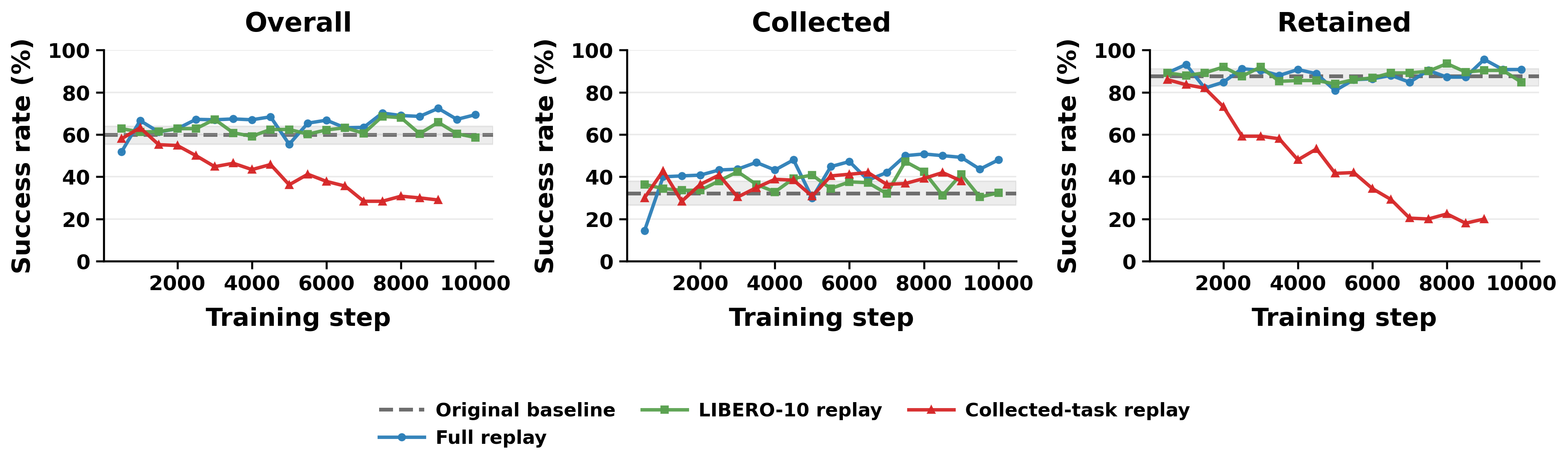}
  \end{minipage}
  \hfill
  \begin{minipage}[b]{0.38\textwidth}
    \includegraphics[width=\linewidth]{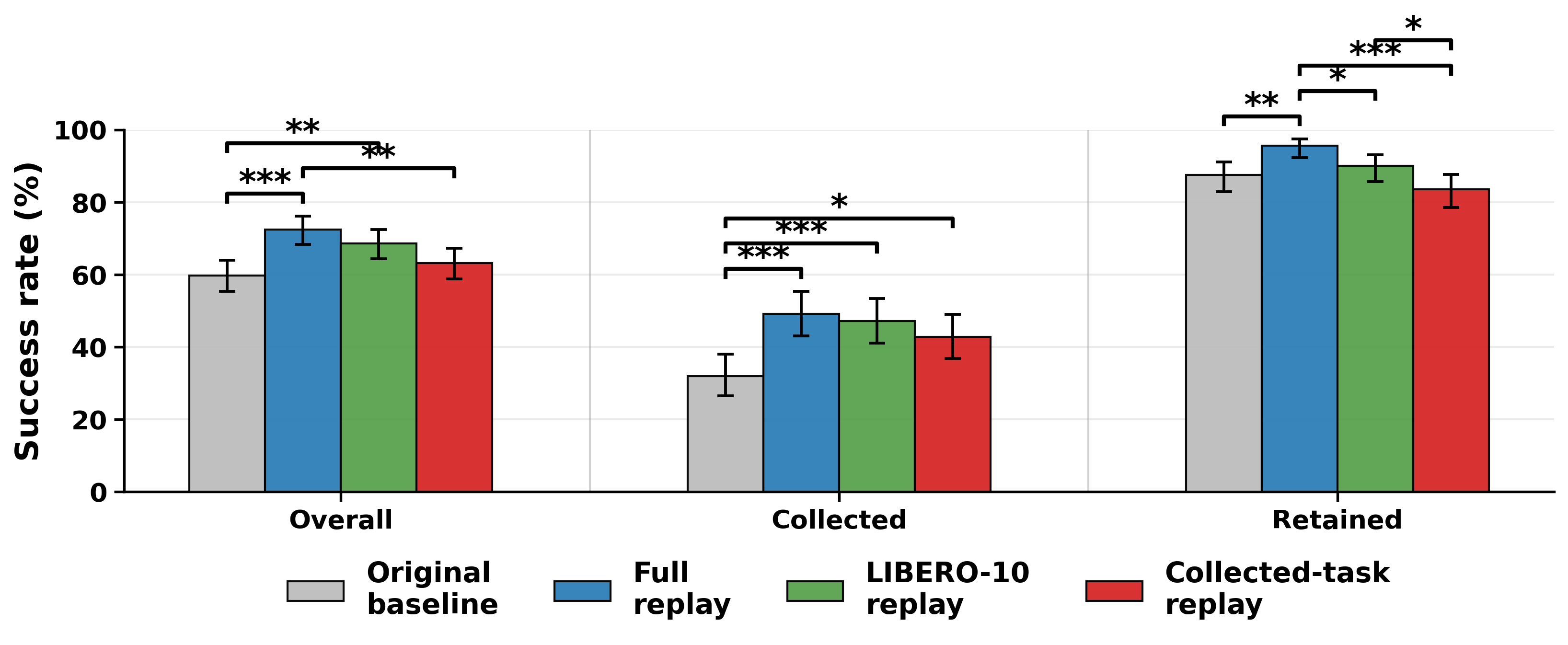}
  \end{minipage}
  \caption{
  \textbf{Replay coverage controls the tradeoff between adaptation and retention.}
  Left: results for Strong INSIGHT online recovery data mixed with different replay subsets. Right: best-checkpoint comparison. 
  }
  \label{fig:rq5_replay_scope}
\end{figure}

\textbf{Results: }
Figure~\ref{fig:rq5_replay_scope} shows that full replay and LIBERO-10 replay both maintain strong retained-task performance while improving collected tasks. LIBERO-10 replay reaches 68.6\% overall success, close to the 72.4\% achieved by full replay; this difference is not significant ($p=0.1877$). In contrast, targeted replay reaches 63.2\% overall success and significantly reduces retained-task performance relative to LIBERO-10 replay, dropping from 90.0\% to 83.6\% retained-task success ($p=0.0345$).

\begin{figure}[t]
    \centering
    \includegraphics[width=\linewidth]{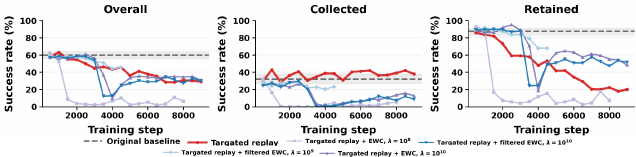}
    \caption{
    \textbf{Regularization does not consistently prevent degradation, indicating that replay coverage is critical.}
    We evaluate EWC and filtered-EWC variants when using targeted replay data. 
    }
    \label{fig:rq5_collected_replay_ewc}
\end{figure}

Figure~\ref{fig:rq5_collected_replay_ewc} further tests whether EWC can stabilize targeted replay. It does not consistently solve the retention problem: some variants briefly preserve performance early in training, but many later degrade substantially. Thus, when replay coverage is too narrow, parameter regularization cannot reliably compensate for missing retained-task data.

\textbf{Discussion: }
Replay does not necessarily need to include the full original dataset, but it must cover the behaviors that should be retained. Restricting replay to LIBERO-10 preserves most of the benefit of full replay, while targeted replay biases adaptation toward the collected tasks and allows non-collected task performance to degrade.

\section{Summary and Conclusion}
\label{sec:results_summary}

We studied active continual learning for autoregressive VLAs by using INSIGHT predictions to identify high-uncertainty, states and collect recovery demonstrations from those states. Across experiments, five findings emerge. First, high-uncertainty recovery demonstrations are more informative than matched passive start-state demonstrations in both the Strong INSIGHT and the Weak INSIGHT settings. Second, online recovery collection is more demonstration-efficient than offline collection: dense offline collection does not provide a statistically significant overall advantage despite requiring more demonstrations. Third, recovery data alone is insufficient for fine-tuning; new-only training causes catastrophic forgetting even under old normalization statistics, which partially constrain distribution shift. Fourth, low learning rates and EWC reduce forgetting but expose a stability-plasticity tradeoff, preserving old behavior at the cost of reduced adaptation. Fifth, replay-based mixing provides the strongest overall solution, but replay coverage matters: LIBERO-10 replay approaches full replay, while targeted replay fails to preserve retained-task performance.

Together, these results support the central claim that active learning for VLAs must be formulated as active continual learning, where robots not only detect when they need help, but also use those moments to collect informative experience and integrate it without erasing prior skills. INSIGHT-style uncertainty predictors identify informative states for recovery collection, but the resulting data is narrow and distribution-shifted. Effective adaptation therefore requires both targeted data collection and mechanisms for preserving previously learned behaviors. 




\section{Limitations}
\label{sec:limitations}

Our study has several limitations. First, all experiments are in LIBERO-10 simulation. This enables controlled evaluation of active data collection and continual adaptation, but physical deployment introduces sensor noise, execution variability, safety constraints, imperfect resets, and higher demonstration costs. Our recovery-collection procedure also assumes access to simulator resets at high-uncertainty intermediate states; on real robots, analogous data would likely require runtime human intervention or demonstrations from naturally reached failure-adjacent states. Thus, the online setting is closer to deployment than offline collection from all high-uncertainty states, but both abstract away real-world collection costs. Second, our results depend on the quality of the help predictor. INSIGHT can produce false positives, which waste demonstrations, and false negatives, which miss useful recovery states; future work should study better calibrated and more risk-sensitive uncertainty estimates. Third, we evaluate one autoregressive VLA family on one benchmark. Other VLA architectures, especially diffusion-based or hybrid action heads, may require different uncertainty signals and recovery-state selection mechanisms. Fourth, our continual-learning methods are simple: replay, low-learning-rate fine-tuning, and EWC. More sophisticated methods such as LoRA, adapters, selective freezing, gradient projection, data reweighting, rehearsal buffers, or distillation may improve the stability-plasticity tradeoff. Finally, we focus primarily on task success; future evaluations should also measure intervention cost, demonstration count, recovery length, action smoothness, safety violations, uncertainty calibration, and repeated active-learning cycles.



\clearpage
\acknowledgments{If a paper is accepted, the final camera-ready version will (and probably should) include acknowledgments. All acknowledgments go at the end of the paper, including thanks to reviewers who gave useful comments, to colleagues who contributed to the ideas, and to funding agencies and corporate sponsors that provided financial support.}


\bibliography{example}  

\newpage
\appendix
\section{Compute Infrastructure}
\label{app:compute_infrastructure}

All models were trained on an institutional high-performance computing cluster using NVIDIA H200 GPUs. Most evaluations were also run on the same cluster, with some additional local testing on an NVIDIA RTX 6000 Ada workstation. In total, the project used approximately 2{,}154 H200 GPU-hours of cluster compute.

\section{Normalization-Statistics Ablations}
\label{app:norm_stats_ablation}

The main active-versus-passive comparison uses new normalization statistics, which better match the state and action distribution of the aggregated replay-plus-recovery dataset. Here, we report additional active-versus-passive comparisons using old normalization statistics. These runs are useful because old normalization statistics partially constrain the fine-tuned policy toward the original data distribution. As a result, old normalization can act like an implicit regularizer: it can help preserve prior behavior, but it can also limit adaptation to the newly collected recovery demonstrations.

\begin{figure}[h]
    \centering
    \includegraphics[width=\linewidth]{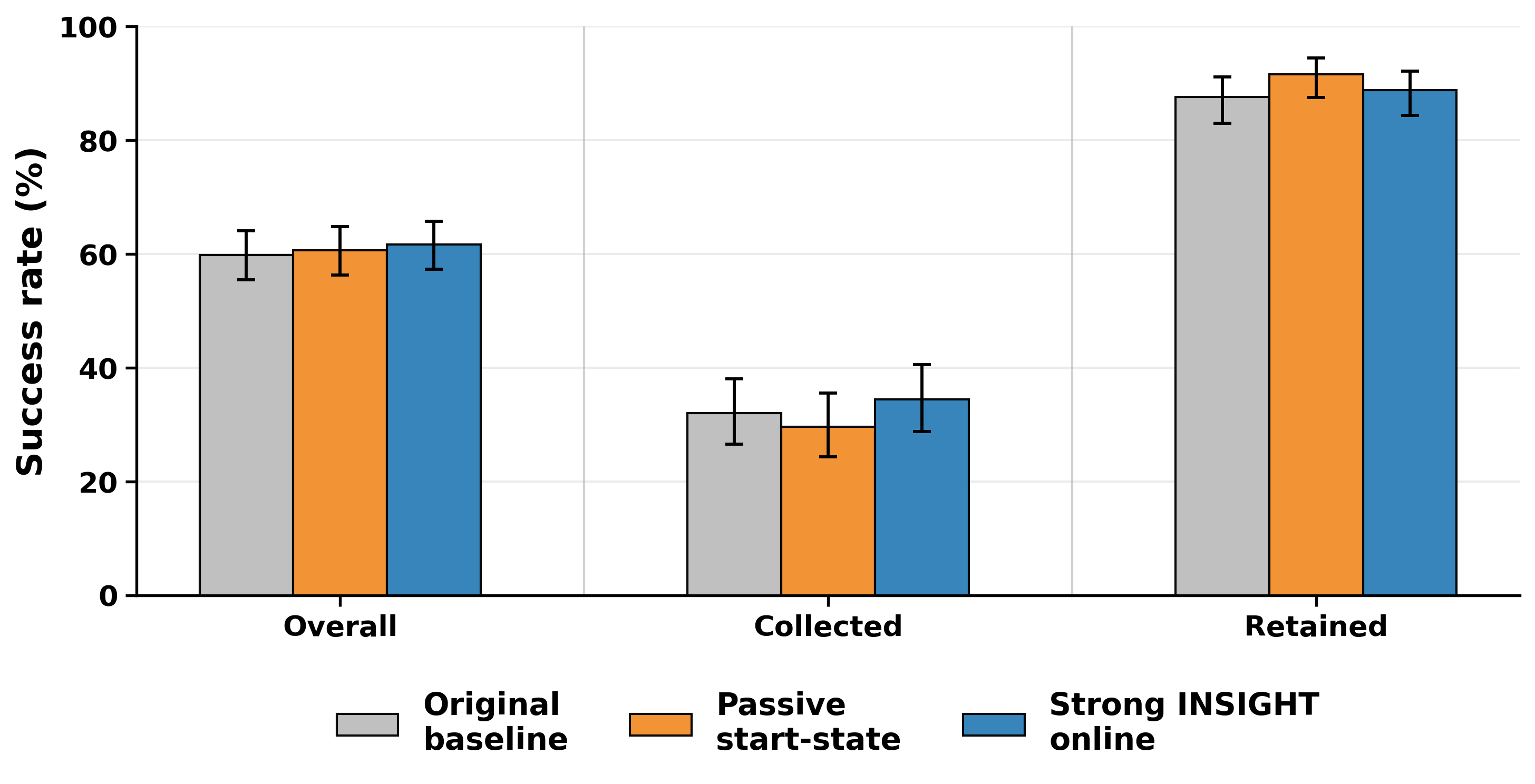}
    \caption{
    \textbf{Strong INSIGHT active-versus-passive comparison with old normalization statistics.}
    Old normalization statistics partially constrain the adapted policy toward the original data distribution. This can preserve retained-task behavior, but it also limits the improvement obtained from high-uncertainty recovery demonstrations.
    }
    \label{fig:app_rq1_active_vs_passive_strong_old_norm}
\end{figure}

Figure~\ref{fig:app_rq1_active_vs_passive_strong_old_norm} shows the Strong INSIGHT active-versus-passive comparison with old normalization statistics. Compared to the main new-normalization result in Figure~\ref{fig:rq1_active_vs_passive_strong}, the improvement from online recovery collection is more muted. This supports the interpretation that old normalization statistics stabilize the model but reduce the plasticity needed to exploit recovery data.

\begin{figure}[h]
    \centering
    \includegraphics[width=\linewidth]{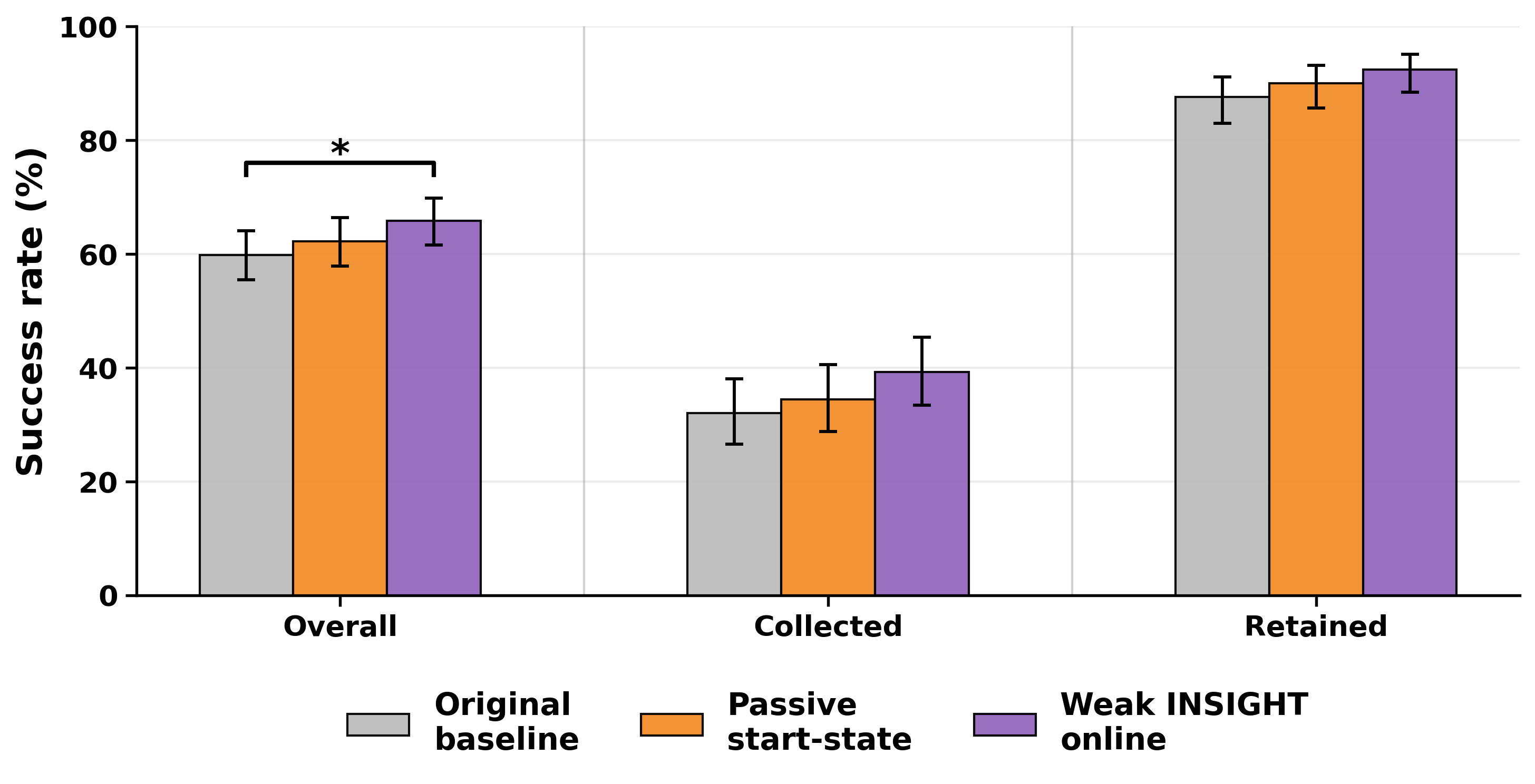}
    \caption{
    \textbf{Weak INSIGHT active-versus-passive comparison with old normalization statistics.}
    This comparison complements the Strong INSIGHT old-normalization ablation and shows that old normalization statistics similarly constrain adaptation under the Weak INSIGHT recovery dataset.
    }
    \label{fig:app_rq1_active_vs_passive_weak_old_norm}
\end{figure}

Figure~\ref{fig:app_rq1_active_vs_passive_weak_old_norm} shows the same old-normalization comparison using Weak INSIGHT. Weak INSIGHT provides a complementary in-domain or LIBERO-aligned help signal, while Strong INSIGHT is the cleaner transfer setting used in the main paper. The old-normalization weak result follows the same qualitative pattern: recovery data can improve collected-task performance, but old normalization limits the magnitude of adaptation.

These ablations also motivate our use of old normalization statistics in the new-only fine-tuning experiments. Since old normalization acts as a conservative stabilizing choice, catastrophic forgetting under old normalization provides stronger evidence that recovery data alone is insufficient for stable adaptation.

\section{Additional EWC Sweeps}
\label{app:ewc_sweeps}

\begin{figure}[t]
    \centering
    \includegraphics[width=\linewidth]{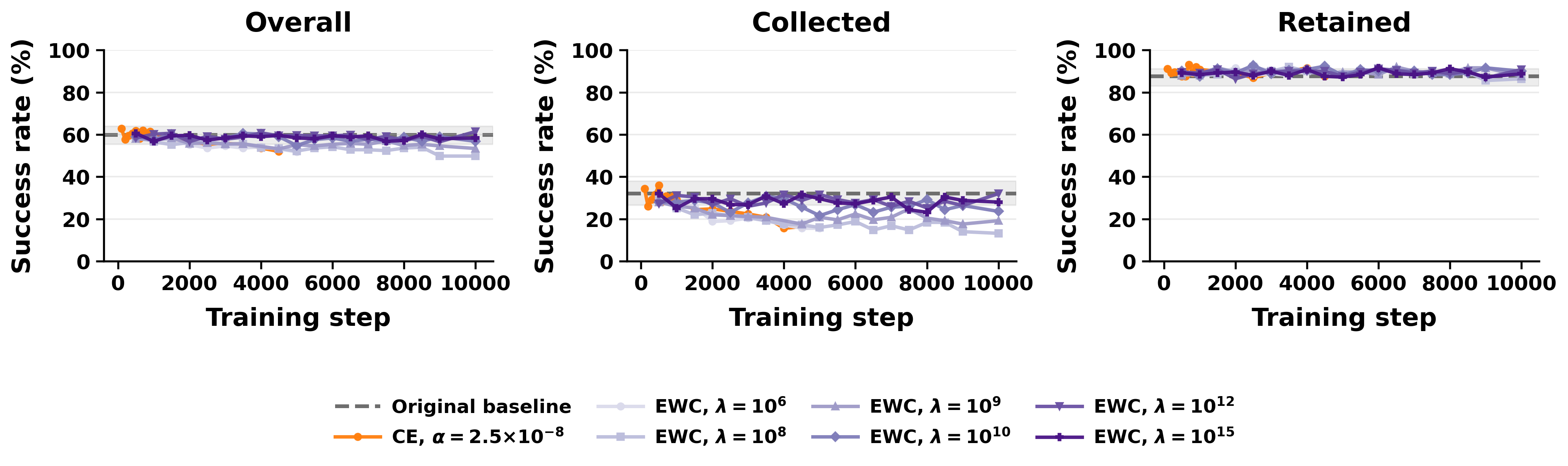}
    \caption{
    \textbf{Full EWC coefficient sweep with low learning rate.}
    Stronger regularization stabilizes retained-task performance, but does not provide enough plasticity to fully exploit uncertainty-guided recovery data.
    }
    \label{fig:app_rq4_ewc_low_lr_full_sweep}
\end{figure}

\begin{figure}[t]
    \centering
    \includegraphics[width=\linewidth]{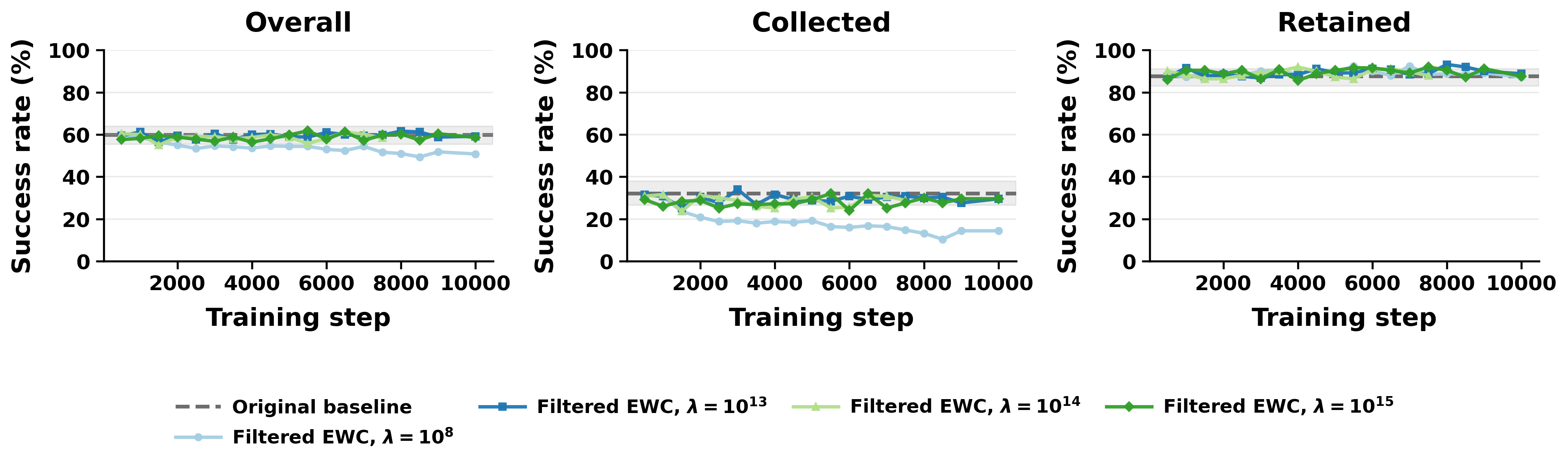}
    \caption{
    \textbf{Filtered Fisher EWC sweep.}
    Computing Fisher information on filtered prior data changes the stability-plasticity behavior, but regularization alone still does not match replay-based adaptation.
    }
    \label{fig:app_rq4_filtered_fisher_sweep}
\end{figure}

Figure~\ref{fig:app_rq4_ewc_low_lr_full_sweep} reports the full EWC coefficient sweep using the low learning rate $\alpha=2.5\times10^{-8}$. Across coefficients, stronger regularization generally improves retention, but collected-task performance remains limited and overall success does not match replay-based mixing. Figure~\ref{fig:app_rq4_filtered_fisher_sweep} reports the corresponding sweep using filtered Fisher estimates. Filtering the Fisher reference data changes the detailed training dynamics, but does not remove the stability-plasticity tradeoff observed in the main paper.

\section{Additional Experimental Results}
\label{app:additional_results}

This appendix reports the best-checkpoint summary, pairwise statistical tests, and raw per-checkpoint task success counts. Best checkpoints are selected by highest overall success over evaluated checkpoints for each method. Full per-checkpoint results are included to make checkpoint selection transparent.

\subsection{Best Checkpoint Summary}
\label{app:best_checkpoint_summary}
\begin{table}[h]
\centering
\small
\caption{\textbf{Best-checkpoint summary for key experimental conditions.} We report overall success, collected-task success, and retained-task success. Each LIBERO-10 task is evaluated with 50 rollouts. Collected tasks are tasks 2, 3, 5, 8, and 9; retained tasks are tasks 0, 1, 4, 6, and 7.}
\label{tab:best_checkpoint_summary}
\resizebox{\linewidth}{!}{
\begin{tabular}{lccccccc}
\toprule
Experiment & Step & Overall & Overall \% & Collected & Collected \% & Retained & Retained \% \\
\midrule
Original baseline & 0 & 299/500 & 59.8 & 80/250 & 32.0 & 219/250 & 87.6 \\
Strong First All-Mix New Norm & 9000 & 362/500 & 72.4 & 123/250 & 49.2 & 239/250 & 95.6 \\
Strong First Baseline All-Mix New Norm & 3000 & 301/500 & 60.2 & 83/250 & 33.2 & 218/250 & 87.2 \\
Strong All All-Mix New Norm & 5000 & 342/500 & 68.4 & 121/250 & 48.4 & 221/250 & 88.4 \\
Weak First All-Mix New Norm & 6000 & 354/500 & 70.8 & 134/250 & 53.6 & 220/250 & 88.0 \\
Weak First All-Mix Old Norm & 500 & 329/500 & 65.8 & 98/250 & 39.2 & 231/250 & 92.4 \\
Weak First Baseline All-Mix Old Norm & 1000 & 311/500 & 62.2 & 86/250 & 34.4 & 225/250 & 90.0 \\
Strong First Only Old Norm & 1000 & 142/500 & 28.4 & 1/250 & 0.4 & 141/250 & 56.4 \\
Strong All Only Old Norm & 1000 & 81/500 & 16.2 & 0/250 & 0.0 & 81/250 & 32.4 \\
Weak First Only Old Norm & 500 & 195/500 & 39.0 & 10/250 & 4.0 & 185/250 & 74.0 \\
Strong First Only Low LR & 100 & 314/500 & 62.8 & 86/250 & 34.4 & 228/250 & 91.2 \\
Fisher 1e12 Low LR & 9999 & 307/500 & 61.4 & 80/250 & 32.0 & 227/250 & 90.8 \\
Strong First Libero 10 Mix & 7500 & 343/500 & 68.6 & 118/250 & 47.2 & 225/250 & 90.0 \\
Strong First Libero10 Collected Mix & 1000 & 316/500 & 63.2 & 107/250 & 42.8 & 209/250 & 83.6 \\
\bottomrule
\end{tabular}
}
\end{table}

\subsection{Pairwise Statistical Tests}
\label{app:pairwise_tests}

\small
\setlength{\tabcolsep}{3pt}
\renewcommand{\arraystretch}{1.08}
\begin{longtable}{
    p{0.55\linewidth}
    p{0.1\linewidth}
    p{0.1\linewidth}
    p{0.1\linewidth}
    p{0.05\linewidth}
    p{0.1\linewidth}
}
\label{tab:pairwise_tests}
\\
\caption{\textbf{Pairwise statistical comparisons for key claims.} We use two-proportion z-tests over rollout success counts. Significance markers denote $^{*}p<0.05$, $^{**}p<0.01$, $^{***}p<0.001$, and n.s. otherwise.}
\label{tab:pairwise_tests}
\\
\toprule
Comparison & Metric & Method A & Method B & $z$ & $p$ \\
\midrule
\endfirsthead
\caption[]{\textbf{Pairwise statistical comparisons for key claims} continued.}
\\
\toprule
Comparison & Metric & Method A & Method B & $z$ & $p$ \\
\midrule
\endhead
\midrule
\multicolumn{6}{r}{\emph{Continued on next page}} \\
\endfoot
\bottomrule
\endlastfoot
RQ1 strong new norm active vs passive overall & Overall & 362/500 & 301/500 & 4.08 & 0.0000$^{***}$ \\
RQ1 strong new norm active vs passive collected & Collected & 123/250 & 83/250 & 3.63 & 0.0003$^{***}$ \\
RQ1 strong new norm active vs passive retained & Retained & 239/250 & 218/250 & 3.35 & 0.0008$^{***}$ \\
RQ1 strong new norm passive vs baseline overall & Overall & 301/500 & 299/500 & 0.13 & 0.8973 n.s. \\
RQ1 strong new norm passive vs baseline collected & Collected & 83/250 & 80/250 & 0.29 & 0.7747 n.s. \\
RQ1 strong new norm passive vs baseline retained & Retained & 218/250 & 219/250 & -0.13 & 0.8928 n.s. \\
RQ1 strong new norm active vs baseline overall & Overall & 362/500 & 299/500 & 4.21 & 0.0000$^{***}$ \\
RQ1 strong new norm active vs baseline collected & Collected & 123/250 & 80/250 & 3.92 & 0.0001$^{***}$ \\
\end{longtable}
\normalsize

\subsection{Raw Per-Checkpoint Results}
\label{app:raw_checkpoint_results}
Tables below report raw task success counts. T0--T9 denote LIBERO-10 tasks. Each task is evaluated with 50 rollouts.

\begin{landscape}
\subsubsection{Active Collection and Replay Baselines}
\small
\begin{longtable}{llrrrrrrrrrrccc}
\caption{Raw per-checkpoint results for Active Collection and Replay Baselines.}
\label{tab:raw_active_collection_and_replay_baselines} \\
\toprule
Experiment & Step & T0 & T1 & T2 & T3 & T4 & T5 & T6 & T7 & T8 & T9 & Overall \% & Collected \% & Retained \% \\
\midrule
\endfirsthead
\toprule
Experiment & Step & T0 & T1 & T2 & T3 & T4 & T5 & T6 & T7 & T8 & T9 & Overall \% & Collected \% & Retained \% \\
\midrule
\endhead
Strong All All-Mix New Norm & 1000 & 0 & 43 & 2 & 20 & 34 & 24 & 30 & 46 & 0 & 3 & 40.4 & 19.6 & 61.2 \\
Strong All All-Mix New Norm & 3000 & 42 & 47 & 6 & 34 & 45 & 46 & 34 & 41 & 1 & 21 & 63.4 & 43.2 & 83.6 \\
Strong All All-Mix New Norm & 5000 & 41 & 45 & 17 & 38 & 42 & 45 & 46 & 47 & 10 & 11 & 68.4 & 48.4 & 88.4 \\
Strong All All-Mix New Norm & 7000 & 45 & 47 & 22 & 41 & 41 & 45 & 30 & 50 & 3 & 14 & 67.6 & 50.0 & 85.2 \\
Strong All All-Mix Old Norm & 1000 & 42 & 46 & 11 & 25 & 46 & 20 & 36 & 48 & 0 & 10 & 56.8 & 26.4 & 87.2 \\
Strong All All-Mix Old Norm & 3000 & 47 & 48 & 3 & 21 & 48 & 37 & 39 & 45 & 0 & 6 & 58.8 & 26.8 & 90.8 \\
Strong All All-Mix Old Norm & 5000 & 36 & 50 & 7 & 18 & 39 & 19 & 43 & 47 & 0 & 1 & 52.0 & 18.0 & 86.0 \\
Strong First All-Mix New Norm & 500 & 45 & 48 & 0 & 15 & 43 & 19 & 41 & 46 & 0 & 2 & 51.8 & 14.4 & 89.2 \\
\bottomrule
\end{longtable}
\end{landscape}

\begin{landscape}
\subsubsection{New-Only Fine-Tuning}
\small
\begin{longtable}{llrrrrrrrrrrccc}
\caption{Raw per-checkpoint results for New-Only Fine-Tuning.}
\label{tab:raw_new_only_fine_tuning} \\
\toprule
Experiment & Step & T0 & T1 & T2 & T3 & T4 & T5 & T6 & T7 & T8 & T9 & Overall \% & Collected \% & Retained \% \\
\midrule
\endfirsthead
\toprule
Experiment & Step & T0 & T1 & T2 & T3 & T4 & T5 & T6 & T7 & T8 & T9 & Overall \% & Collected \% & Retained \% \\
\midrule
\endhead
Strong All Only New Norm & 1000 & 0 & 0 & 0 & 0 & 0 & 0 & 0 & 0 & 0 & 0 & 0.0 & 0.0 & 0.0 \\
Strong All Only New Norm & 3000 & 0 & 0 & 0 & 0 & 0 & 0 & 0 & 0 & 0 & 0 & 0.0 & 0.0 & 0.0 \\
Strong All Only New Norm & 5000 & 0 & 0 & 0 & 0 & 0 & 0 & 0 & 0 & 0 & 0 & 0.0 & 0.0 & 0.0 \\
Strong All Only Old Norm & 1000 & 15 & 7 & 0 & 0 & 11 & 0 & 18 & 30 & 0 & 0 & 16.2 & 0.0 & 32.4 \\
Strong All Only Old Norm & 3000 & 1 & 0 & 0 & 0 & 0 & 0 & 0 & 3 & 0 & 0 & 0.8 & 0.0 & 1.6 \\
Strong All Only Old Norm & 5000 & 0 & 2 & 0 & 0 & 0 & 0 & 0 & 0 & 0 & 0 & 0.4 & 0.0 & 0.8 \\
Strong First Only New Norm & 1000 & 0 & 0 & 0 & 0 & 0 & 0 & 0 & 0 & 0 & 0 & 0.0 & 0.0 & 0.0 \\
Strong First Only New Norm & 3000 & 0 & 0 & 0 & 0 & 0 & 0 & 0 & 0 & 0 & 0 & 0.0 & 0.0 & 0.0 \\
\bottomrule
\end{longtable}
\end{landscape}

\begin{landscape}
\subsubsection{Low-Learning-Rate and EWC Sweeps}
\small
\begin{longtable}{llrrrrrrrrrrccc}
\caption{Raw per-checkpoint results for Low-Learning-Rate and EWC Sweeps.}
\label{tab:raw_low_learning_rate_and_ewc_sweeps} \\
\toprule
Experiment & Step & T0 & T1 & T2 & T3 & T4 & T5 & T6 & T7 & T8 & T9 & Overall \% & Collected \% & Retained \% \\
\midrule
\endfirsthead
\toprule
Experiment & Step & T0 & T1 & T2 & T3 & T4 & T5 & T6 & T7 & T8 & T9 & Overall \% & Collected \% & Retained \% \\
\midrule
\endhead
Filtered Fisher 1e13 Low LR & 500 & 44 & 49 & 19 & 21 & 41 & 25 & 38 & 46 & 0 & 14 & 59.4 & 31.6 & 87.2 \\
Filtered Fisher 1e13 Low LR & 1000 & 44 & 48 & 14 & 21 & 47 & 29 & 43 & 47 & 0 & 13 & 61.2 & 30.8 & 91.6 \\
Filtered Fisher 1e13 Low LR & 1500 & 42 & 49 & 14 & 16 & 46 & 24 & 37 & 45 & 0 & 9 & 56.4 & 25.2 & 87.6 \\
Filtered Fisher 1e13 Low LR & 2000 & 43 & 49 & 17 & 18 & 44 & 28 & 40 & 45 & 0 & 13 & 59.4 & 30.4 & 88.4 \\
Filtered Fisher 1e13 Low LR & 2500 & 41 & 48 & 13 & 17 & 45 & 22 & 41 & 44 & 0 & 17 & 57.6 & 27.6 & 87.6 \\
Filtered Fisher 1e13 Low LR & 3000 & 40 & 48 & 18 & 18 & 46 & 26 & 36 & 47 & 2 & 21 & 60.4 & 34.0 & 86.8 \\
Filtered Fisher 1e13 Low LR & 3500 & 42 & 49 & 19 & 14 & 43 & 22 & 41 & 46 & 0 & 12 & 57.6 & 26.8 & 88.4 \\
Filtered Fisher 1e13 Low LR & 4000 & 45 & 49 & 16 & 16 & 46 & 29 & 37 & 44 & 0 & 18 & 60.0 & 31.6 & 88.4 \\
\bottomrule
\end{longtable}
\end{landscape}

\begin{landscape}
\subsubsection{LIBERO-10 Replay Scope}
\small
\begin{longtable}{llrrrrrrrrrrccc}
\caption{Raw per-checkpoint results for LIBERO-10 Replay Scope.}
\label{tab:raw_libero_10_replay_scope} \\
\toprule
Experiment & Step & T0 & T1 & T2 & T3 & T4 & T5 & T6 & T7 & T8 & T9 & Overall \% & Collected \% & Retained \% \\
\midrule
\endfirsthead
\toprule
Experiment & Step & T0 & T1 & T2 & T3 & T4 & T5 & T6 & T7 & T8 & T9 & Overall \% & Collected \% & Retained \% \\
\midrule
\endhead
Strong First Libero 10 Mix & 500 & 45 & 49 & 21 & 14 & 46 & 31 & 35 & 48 & 1 & 24 & 62.8 & 36.4 & 89.2 \\
Strong First Libero 10 Mix & 1000 & 43 & 47 & 22 & 12 & 42 & 29 & 40 & 48 & 0 & 23 & 61.2 & 34.4 & 88.0 \\
Strong First Libero 10 Mix & 1500 & 41 & 48 & 20 & 21 & 44 & 28 & 44 & 46 & 0 & 15 & 61.4 & 33.6 & 89.2 \\
Strong First Libero 10 Mix & 2000 & 45 & 49 & 16 & 20 & 44 & 29 & 43 & 49 & 0 & 19 & 62.8 & 33.6 & 92.0 \\
Strong First Libero 10 Mix & 2500 & 44 & 47 & 16 & 27 & 44 & 31 & 41 & 43 & 2 & 19 & 62.8 & 38.0 & 87.6 \\
Strong First Libero 10 Mix & 3000 & 45 & 49 & 25 & 29 & 46 & 34 & 42 & 48 & 1 & 17 & 67.2 & 42.4 & 92.0 \\
Strong First Libero 10 Mix & 3500 & 45 & 46 & 21 & 19 & 43 & 25 & 34 & 45 & 0 & 26 & 60.8 & 36.4 & 85.2 \\
Strong First Libero 10 Mix & 4000 & 42 & 47 & 24 & 21 & 45 & 24 & 34 & 46 & 0 & 13 & 59.2 & 32.8 & 85.6 \\
\bottomrule
\end{longtable}
\end{landscape}

\begin{landscape}
\subsubsection{Collected-Task Replay and EWC}
\small
\begin{longtable}{llrrrrrrrrrrccc}
\caption{Raw per-checkpoint results for Collected-Task Replay and EWC.}
\label{tab:raw_collected_task_replay_and_ewc} \\
\toprule
Experiment & Step & T0 & T1 & T2 & T3 & T4 & T5 & T6 & T7 & T8 & T9 & Overall \% & Collected \% & Retained \% \\
\midrule
\endfirsthead
\toprule
Experiment & Step & T0 & T1 & T2 & T3 & T4 & T5 & T6 & T7 & T8 & T9 & Overall \% & Collected \% & Retained \% \\
\midrule
\endhead
Strong First Libero10 Collected Mix & 500 & 44 & 46 & 18 & 18 & 41 & 23 & 37 & 47 & 0 & 16 & 58.0 & 30.0 & 86.0 \\
Strong First Libero10 Collected Mix & 1000 & 44 & 39 & 24 & 31 & 43 & 33 & 35 & 48 & 0 & 19 & 63.2 & 42.8 & 83.6 \\
Strong First Libero10 Collected Mix & 1500 & 47 & 32 & 15 & 18 & 41 & 24 & 42 & 43 & 1 & 13 & 55.2 & 28.4 & 82.0 \\
Strong First Libero10 Collected Mix & 2000 & 47 & 24 & 19 & 22 & 31 & 32 & 38 & 43 & 1 & 17 & 54.8 & 36.4 & 73.2 \\
Strong First Libero10 Collected Mix & 2500 & 46 & 20 & 20 & 30 & 10 & 34 & 35 & 37 & 2 & 16 & 50.0 & 40.8 & 59.2 \\
Strong First Libero10 Collected Mix & 3000 & 42 & 29 & 17 & 26 & 10 & 18 & 28 & 39 & 0 & 15 & 44.8 & 30.4 & 59.2 \\
Strong First Libero10 Collected Mix & 3500 & 48 & 28 & 14 & 28 & 8 & 27 & 30 & 31 & 2 & 16 & 46.4 & 34.8 & 58.0 \\
Strong First Libero10 Collected Mix & 4000 & 35 & 26 & 16 & 24 & 3 & 36 & 19 & 37 & 3 & 18 & 43.4 & 38.8 & 48.0 \\
\bottomrule
\end{longtable}
\end{landscape}

\end{document}